\documentclass[letterpaper]{llncs}
\usepackage{times}
\usepackage{helvet}
\usepackage{courier}
\usepackage{epic}
\usepackage{graphicx}
\usepackage{latexsym}
\usepackage{verbatim}
\usepackage{xspace}
\begin{document}

\newlength{\halftextwidth}
\setlength{\halftextwidth}{0.47\textwidth}
\def\halffigsize{2.2in}
\def\thirdfigsize{1.5in}
\def\negvspace{0in}
\def\posvspace{0em}






\newcommand{\set}{\mathcal}
\newcommand{\myset}[1]{\ensuremath{\mathcal #1}}

\renewcommand{\theenumii}{\alph{enumii}}
\renewcommand{\theenumiii}{\roman{enumiii}}
\newcommand{\figref}[1]{Figure \ref{#1}}
\newcommand{\tref}[1]{Table \ref{#1}}
\newcommand{\myOmit}[1]{}
\newcommand{\And}{\wedge}
\newcommand{\myldots}{.}

\newcommand{\nina}[1]{{#1}}

\newtheorem{mydefinition}{Definition}
\newtheorem{mytheorem}{Theorem}
\newtheorem{mycor}{Corollary}
\newenvironment{myexample}{{\bf Running example:} \it}{\rm}
\newtheorem{mytheorem1}{Theorem}
\newcommand{\myproof}{\noindent {\bf Proof:\ \ }}
\newcommand{\myqed}{\mbox{$\Box$}}

\newcommand{\mymod}{\mbox{\rm mod}}
\newcommand{\range}{\mbox{\sc Range}}
\newcommand{\roots}{\mbox{\sc Roots}}
\newcommand{\myiff}{\mbox{\rm iff}}
\newcommand{\alldifferent}{\mbox{\sc AllDifferent}}
\newcommand{\alldiff}{\mbox{\sc AllDifferent}}
\newcommand{\interdistance}{\mbox{\sc InterDistance}}
\newcommand{\permutation}{\mbox{\sc Permutation}}
\newcommand{\disjoint}{\mbox{\sc Disjoint}}
\newcommand{\cardpath}{\mbox{\sc CardPath}}
\newcommand{\CARDPATH}{\mbox{\sc CardPath}}
\newcommand{\knapsack}{\mbox{\sc Knapsack}}
\newcommand{\common}{\mbox{\sc Common}}
\newcommand{\uses}{\mbox{\sc Uses}}
\newcommand{\lex}{\mbox{\sc Lex}}
\newcommand{\LEX}{\mbox{\sc Lex}}
\newcommand{\SnakeLex}{\mbox{\sc SnakeLex}}
\newcommand{\usedby}{\mbox{\sc UsedBy}}
\newcommand{\nvalue}{\mbox{\sc NValue}}
\newcommand{\slide}{\mbox{\sc Slide}}
\newcommand{\SLIDE}{\mbox{\sc Slide}}
\newcommand{\circularslide}{\mbox{\sc Slide}_{\rm O}}
\newcommand{\among}{\mbox{\sc Among}}
\newcommand{\mysum}{\mbox{\sc Sum}}
\newcommand{\amongseq}{\mbox{\sc AmongSeq}}
\newcommand{\atmost}{\mbox{\sc AtMost}}
\newcommand{\atleast}{\mbox{\sc AtLeast}}
\newcommand{\element}{\mbox{\sc Element}}
\newcommand{\gcc}{\mbox{\sc Gcc}}
\newcommand{\egcc}{\mbox{\sc EGcc}}
\newcommand{\gsc}{\mbox{\sc Gsc}}
\newcommand{\contiguity}{\mbox{\sc Contiguity}}
\newcommand{\PRECEDENCE}{\mbox{\sc Precedence}}
\newcommand{\precedence}{\mbox{\sc Precedence}}
\newcommand{\assignnvalues}{\mbox{\sc Assign\&NValues}}
\newcommand{\linksettobooleans}{\mbox{\sc LinkSet2Booleans}}
\newcommand{\domain}{\mbox{\sc Domain}}
\newcommand{\symalldiff}{\mbox{\sc SymAllDiff}}
\newcommand{\valsymbreak}{\mbox{\sc ValSymBreak}}
\newcommand{\RowColSym}{\mbox{\sc RowColLexLeader}}
\newcommand{\RowColSymShort}{\mbox{\sc RowColLex}}
\newcommand{\RowSymShort}{\mbox{\sc RowLex}}
\newcommand{\LexLeader}{\mbox{\sc LexLeader}}
\newcommand{\ColSymShort}{\mbox{\sc ColLex}}
\newcommand{\NoSymShort}{\mbox{\sc NoSB}}
\newcommand{\RowLexLeader}{\mbox{\sc RowLexLeader}}
\newcommand{\LexChain}{\mbox{\sc LexChain}}
\newcommand{\OrderRowCol}{\mbox{\sc Order1stRowCol}}

\newcommand{\slidingsum}{\mbox{\sc SlidingSum}}
\newcommand{\MaxIndex}{\mbox{\sc MaxIndex}}
\newcommand{\REGULAR}{\mbox{\sc Regular}}
\newcommand{\regular}{\mbox{\sc Regular}}
\newcommand{\Regular}{\mbox{\sc Regular}}
\newcommand{\STRETCH}{\mbox{\sc Stretch}}
\newcommand{\SLIDEOR}{\mbox{\sc SlideOr}}
\newcommand{\NAE}{\mbox{\sc NotAllEqual}}
\newcommand{\mymax}{\mbox{\rm max}}

\newcommand{\todo}[1]{{\tt (... #1 ...)}}

\newcommand{\DC}{\ensuremath{DC}\xspace}
\newcommand{\Xbf}{\mbox{{\bf X}}\xspace}
\newcommand{\LEXCHAIN}{\mbox{\sc LexChain}}
\newcommand{\DLex}{\mbox{\sc DoubleLex}\xspace}
\newcommand{\snakelex}{\mbox{\sc SnakeLex}\xspace}
\newcommand{\DLexColSum}{\mbox{\sc DoubleLexColSum}\xspace}

\title{Symmetry within and between solutions\thanks{
Supported by the Australian 
Government's  Department of Broadband, Communications and the Digital Economy
and the
ARC. Thanks to the co-authors of the work summarized here: Marijn Heule, George Katsirelos and 
Nina Narodytska.}}
\author{
Toby Walsh}
\institute{NICTA and University of NSW,
Sydney, Australia, email: 
toby.walsh@nicta.com.au
}

\maketitle
\begin{abstract}
Symmetry can be used to help solve many problems. For instance,
Einstein's famous 1905 paper ("On the Electrodynamics of Moving Bodies")
uses symmetry to help derive the laws of special relativity.
In artificial intelligence, symmetry has played an important
role in both problem representation and reasoning.
I describe recent work on using symmetry
to help solve constraint satisfaction problems. 
Symmetries occur within individual solutions
of problems as well as between different
solutions of the same problem. 
Symmetry can also be applied to the constraints
in a problem to give new symmetric constraints. 
Reasoning about symmetry
can speed up problem solving, and has led to
the discovery of new results in both graph and number theory.
\end{abstract}

\myOmit{

}

\section{Introduction}

Symmetry occurs in 
many combinatorial search
problems. For example, 
in the 
magic squares problem (prob019 in CSPLib \cite{csplib}),
we have the symmetries that rotate and reflect the square.
Eliminating such symmetry from the search
space is often critical when trying to 
solve large instances of a problem. 
Symmetry can occur both {\em within}
a single solution as well as {\em between}
different solutions of a problem.
We can also {\em apply} symmetry to the constraints
in a problem. 
We focus here on constraint satisfaction
problems, though there has been interesting
work on symmetry in other types of
problems (e.g. planning, and model checking). 
We summarize recent work appearing
in \cite{hwaaai10,kwecai10,knwcp10}.

\section{Symmetry between solutions}

A symmetry $\sigma$ is a bijection on assignments. 
Given a set of assignments $A$
and a symmetry $\sigma$,
we write $\sigma(A)$ for $\{ \sigma(a) \ | \ a \in A\}$. 
A special type of symmetry, called 
\emph{solution symmetry}  
is a symmetry \emph{between} the
solutions of a problem. 
More formally, we say that a problem has 
the \emph{solution symmetry} $\sigma$ iff $\sigma$ of
any solution is itself a solution
\cite{cjjpsconstraints06}. 

\begin{myexample}
The \emph{magic squares} problem is to label a $n$ by $n$
square so that the sum of every row, column
and diagonal are equal (prob019 in CSPLib \cite{csplib}). 
A \emph{normal} magic square contains the integers 1 to $n^2$. 
We model this with $n^2$ variables
$X_{i,j}$ where
$X_{i,j}=k$ iff the $i$th column and $j$th row is 
labelled with the integer $k$. 

``Lo Shu'', the smallest non-trivial
normal magic square has been known for over four thousand years and is 
an important object in ancient Chinese
mathematics:
\begin{eqnarray} \label{loshu}
&
\begin{tabular}{|c|c|c|} \hline
4 & 9 & 2 \\ \hline
3 & 5 & 7 \\ \hline
8 & 1 & 6 \\ \hline
\end{tabular}
&
\end{eqnarray}

The magic squares problem has a number of  solution symmetries. 
For example, consider the symmetry $\sigma_d$
that reflects a solution in the leading diagonal. 
This map ``Lo Shu'' onto a
symmetric solution:
\begin{eqnarray} \label{loshu2}
&
\begin{tabular}{|c|c|c|} \hline
6 & 7 & 2 \\ \hline
1 & 5 & 9 \\ \hline
8 & 3 & 4 \\ \hline
\end{tabular}
&
\end{eqnarray}
Any other rotation or reflection of
the square maps one solution onto
another. The 8 symmetries
of the square 
are thus all solution symmetries 
of this problem. In fact, there are only 8 different magic
square of order 3, and all are in the same symmetry
class. 
\end{myexample}

One way to factor solution symmetry
out of the search space is to post symmetry
breaking constraints. See, for instance,
\cite{puget:Sym,clgrkr96,ffhkmpwcp2002,fhkmwcp2002,fhkmwaij2005,wcp06,wecai2006,llwycp07,wcp07}.
For example, we can eliminate
$\sigma_d$ by posting a constraint which ensures that the 
top left corner is smaller than its symmetry, the bottom right
corner. This selects (\ref{loshu}) and
eliminates (\ref{loshu2}). 
Symmetry can be used to transform
such symmetry breaking constraints \cite{hwaaai10}.
For example, 
if we apply $\sigma_d$ to  the constraint which ensures that the 
top left corner is smaller than the bottom right,
we get a new symmetry breaking constraints
which ensures that the bottom right is smaller
than the top left. This selects (\ref{loshu2}) and
eliminates (\ref{loshu}). 

\section{Symmetry within a solution}

Symmetries can also be found {within} individual solutions 
of a constraint satisfaction problem. 
We say that a solution $A$ \emph{contains}
the internal symmetry $\sigma$ (or equivalently
$\sigma$ is a internal symmetry \emph{within} this solution)
iff $\sigma(A)=A$. 

\begin{myexample}
Consider again 
``Lo Shu''. 
This contains an internal symmetry. 
To see this, 
consider the solution symmetry $\sigma_{inv}$ that
inverts labels, mapping $k$ 
onto $n^2+1-k$. This solution symmetry maps
``Lo Shu'' onto a different (but symmetric) solution.
However, if we now apply the solution symmetry $\sigma_{180}$ that
rotates the square $180^\circ$, we 
map back onto the original solution:
\begin{eqnarray*}
\begin{tabular}{|c|c|c|} \hline
4 & 9 & 2 \\ \hline
3 & 5 & 7 \\ \hline
8 & 1 & 6 \\ \hline
\end{tabular}
& 
\begin{array}{c}
\sigma_{inv} \\
\Rightarrow \\
\Leftarrow \\
\sigma_{180}
\end{array}
&
\begin{tabular}{|c|c|c|} \hline
6 & 1 & 8 \\ \hline
7 & 5 & 3 \\ \hline
2 & 9 & 4 \\ \hline
\end{tabular}
\end{eqnarray*}

Consider 
the composition of these two symmetries:
$\sigma_{inv} \circ \sigma_{180}$. 
As this 
maps ``Lo Shu'' onto itself,
the solution ``Lo Shu'' contains 
the internal symmetry $\sigma_{inv} \circ \sigma_{180}$.
\end{myexample}

In general, there is no relationship
between the solution symmetries of a problem
and the internal symmetries within a solution of
that problem. 
There are solution symmetries of a problem
which are not internal symmetries within any solution
of that problem, and vice versa. 
However, when all solutions of a problem
contain the same internal symmetry, we can be sure that this
is a solution symmetry of the problem itself.
The exploitation of internal symmetries
involves two steps: finding internal symmetries, 
and then restricting search to
solutions containing just these internal symmetries.
%
We have explored this idea in two applications where we
have been able to extend the state of the 
art. In the first, we found new
lower bound certificates for Van der Waerden numbers. 
Such numbers are an important concept in Ramsey theory.
In the second application, we increased the size of 
graceful labellings known for a family of
graphs. Graceful labelling has 
practical applications in 
areas like communication theory. 
Before our work, the largest double wheel graph that we found
graceful labelled in the 
literature had size 
10. 
Using our method, 
we constructed the first known labelling for a double wheel of size 
$24$.

\bibliographystyle{splncs}


\end{document}